\documentclass[sigconf,natbib=false]{acmart}

\setcopyright{acmlicensed}
\copyrightyear{2027}
\acmYear{2027}
\acmDOI{XXXXXXX.XXXXXXX}
\acmConference[SAC'27]{The 42nd ACM/SIGAPP Symposium on Applied Computing}{April 5--9, 2027}{Gwangju, South Korea}
\acmISBN{979-X-XXXX-XXXX-X/27/04}

\RequirePackage[
  datamodel=acmdatamodel,
  style=acmnumeric,
  natbib=true,
  maxbibnames=3,   
  minbibnames=1,
  ]{biblatex}
\AtEveryBibitem{%
  \clearfield{isbn}\clearfield{issn}\clearfield{numpages}\clearfield{month}%
  \clearfield{articleno}\clearfield{eid}\clearfield{issue_date}%
  \clearname{editor}\clearlist{location}\clearlist{address}%
}
\usepackage{multirow}
\usepackage{amsmath}
\usepackage{graphicx}
\usepackage{xcolor}
\usepackage{tcolorbox}
\tcbuselibrary{skins, breakable}
\usepackage{listings}

\newif\ifappendix
\appendixfalse

\begin{document}

\title{SE-MoLoRA: Shared-Expert LoRA Adapters for Domain-Specific Photographic Assessment}
\renewcommand{\shorttitle}{SE-MoLoRA: Shared-Expert LoRA Adapters for Photographic Assessment}

\author{Bishwash Khanal}
\email{bishwash.b.khanal@jyu.fi}
\orcid{0009-0008-6972-4660}
\affiliation{%
  \institution{University of Jyv\"askyl\"a}
  \city{Jyv\"askyl\"a}
  \country{Finland}}

\author{Anlan Zhang}
\authornote{Acted in an advisor capacity only; did not process, store, download, or direct the use of project data or models.}
\email{anlanz@adobe.com}
\orcid{0000-0003-2371-4631}
\affiliation{%
  \institution{Adobe Research}
  \city{San Jose}
  \state{California}
  \country{USA}}

\author{Sasu Tarkoma}
\email{sasu.tarkoma@helsinki.fi}
\orcid{0000-0003-4220-3650}
\affiliation{%
  \institution{University of Helsinki}
  \city{Helsinki}
  \country{Finland}}

\author{Tommi Mikkonen}
\email{tommi.j.mikkonen@jyu.fi}
\orcid{0000-0002-8540-9918}
\affiliation{%
  \institution{University of Jyv\"askyl\"a}
  \city{Jyv\"askyl\"a}
  \country{Finland}}

\author{Abhishek Kumar}
\authornote{Corresponding Author.}
\email{abhishek.k.kumar@jyu.fi}
\orcid{0000-0003-4383-7225}
\affiliation{%
  \institution{University of Jyv\"askyl\"a}
  \city{Jyv\"askyl\"a}
  \country{Finland}}

\renewcommand{\shortauthors}{B. Khanal et al.}

\begin{abstract}

Vision-language models can describe images fluently but often struggle to provide actionable photographic critique because semantic understanding and aesthetic judgment remain entangled. We propose SE-MoLoRA, a modular, parameter-efficient framework for adapting a vision-language model to domain-specific photographic assessment. SE-MoLoRA combines an always-active shared LoRA adapter, which captures cross-domain photographic knowledge, with specialist adapters that model residual expertise in composition, lighting, and technical quality. The specialists are trained over the shared representation using orthogonal regularization, and a multimodal router selects a specialist based on the input image and user query. We evaluate SE-MoLoRA on domain-specific critique data distilled from the Reddit Photo Critique Dataset. Across the shared adapter and three specialists, SE-MoLoRA yields relative improvements over monolithic LoRA of 15--82\% in BERTScore F1, 3--27\% in SBERT cosine similarity, and 16--87\% in ROUGE-L F1. It is preferred over the zero-shot base model in 55--73\% of LLM-judged pairwise comparisons, and the shared adapter is preferred over monolithic LoRA in 84.6\% of comparisons. Finally, SVD-based analysis shows that the mean cosine similarity among specialist adapters decreases from 0.64 to 0.53 when the shared adapter is introduced and to 0.43 with orthogonal regularization, indicating more distinct and complementary specialization. The source code, prompts, and evaluation suite for SE-MoLoRA are publicly available at \url{https://github.com/bikhanal/semolora}.
\end{abstract}



\maketitle

\section{Introduction}


Recent Vision-Language Models (VLMs) have achieved strong general-purpose visual understanding, yet their ability to provide meaningful and actionable aesthetic critiques remains limited~\citep{10.1145/3746027.3754961, wang2025aestest, 10.1145/3664647.3680649}. For example, a model may fluently describe ``a beautiful sunset over the ocean,'' yet fail to tell a photographer whether the horizon is tilted, the composition lacks a clear visual anchor, or the exposure obscures important details. Photographic critique therefore requires more than recognizing \textit{what} appears in an image: it also requires assessing \textit{how effectively} the image has been captured across multiple aesthetic and technical dimensions. A key challenge is that such judgments may be influenced by semantic content, resulting in biases such as the halo effect, whereby visually or semantically appealing subjects receive favorable assessments despite technical deficiencies~\citep{bulat2025beauty}.

A natural way to improve photographic critique is to specialize a VLM for different dimensions of photographic quality, such as composition, lighting, and technical quality. However, these dimensions are not independent: composition, lighting, and technical assessment all rely on shared understanding of the image content, subject, and scene structure. Effective critique therefore requires both \emph{shared photographic knowledge} and \emph{domain-specific expertise}. A monolithic adaptation strategy offers no explicit mechanism for separating these two forms of knowledge, while training independent models for each critique domain redundantly represents their shared knowledge and increases storage and deployment costs as the number of domains grows~\citep{houlsby19a}. Mixture-of-Experts and modular adapter methods provide a natural alternative~\citep{2017olln, mixture, almaamari2025mmoe}, but existing approaches are primarily designed for multi-task or general instruction adaptation rather than semantically overlapping critique domains. This motivates a key design question: \textit{How can a model specialize across photographic critique domains while efficiently sharing the knowledge common to them?}

To address this challenge, we propose \textbf{SE-MoLoRA}, a shared-specialist parameter-efficient adaptation framework for controllable photographic critique. SE-MoLoRA decomposes adaptation into an always-active shared LoRA adapter that captures cross-domain photographic knowledge and a set of routed specialist LoRA adapters that model residual expertise in composition, lighting, and technical quality. Rather than learning each domain independently from the frozen base model, we first adapt the model with the shared expert and then train each specialist to capture domain-specific residuals. Because domain-separated supervision alone does not guarantee specialization, we further apply orthogonal regularization to specialist LoRA updates to reduce overlap among their learned directions. This shared-specialist decomposition represents common knowledge once while preserving modular, domain-specific adaptation.

A second challenge is determining which specialist should be activated for a given request. Explicit queries such as ``How is the lighting?'' provide a clear routing signal, but users may instead ask vague questions such as ``What do you think?'' In such cases, visual cues from the image can provide complementary evidence for identifying a relevant critique domain. We therefore employ a multimodal router that jointly considers the input image and user query to select the appropriate specialist. At inference time, SE-MoLoRA activates the shared adapter and at most one specialist adapter, enabling targeted critique without requiring a separate full model for each domain.

We train and evaluate SE-MoLoRA using domain-specific supervision distilled from the Reddit Photo Critique Dataset~\citep{10.5555/3600270.3602745}. Our experiments compare SE-MoLoRA with the zero-shot base model and monolithic LoRA adaptation using reference-based critique metrics and LLM-judged pairwise comparisons. SE-MoLoRA consistently improves BERTScore F1, SBERT cosine similarity, and ROUGE-L F1 across critique domains, and achieves competitive or improved performance relative to monolithic LoRA in pairwise evaluation. We further analyze the learned LoRA updates and find that the shared-specialist decomposition and orthogonal regularization reduce overlap among specialist adapters, supporting the intended separation between shared and domain-specific expertise.

Our contributions are threefold: \begin{itemize} 

\item We frame domain-specific photographic critique as a \emph{shared-specialist adaptation problem}, recognizing that semantically overlapping critique domains require a combination of shared photographic knowledge and domain-specific residual expertise. 

\item We propose SE-MoLoRA, a parameter-efficient architecture that integrates an always-active shared LoRA adapter with routed residual specialists and applies orthogonal regularization to promote specialization among expert adapters. 

\item We introduce a multimodal router for controllable specialist selection and conduct experiments demonstrating that SE-MoLoRA improves critique alignment over zero-shot and monolithic LoRA baselines. Our analysis further shows that the shared-specialist design facilitates knowledge sharing, while orthogonal regularization reduces overlap among specialist adapters. 
\end{itemize}
\section{Related Work} \label{sec:related-work}

\subsection{Image Aesthetic Assessment and Photographic Critique}

Automated Image Aesthetic Assessment (IAA) has shifted from score prediction~\citep{talebi2018nima} through attribute-based approaches predicting factors like composition, lighting, and depth of field~\citep{10.1145/3343031.3350970}, to multimodal frameworks for critique generation. In parallel, vision-language models (VLMs) have evolved from domain-specific systems into broadly capable multimodal assistants~\citep{10.1561/0600000110, zhang2024vlmsurvey}, integrating strong visual encoders with large language backbones to improve visual question answering, scene interpretation, and reasoning-intensive tasks~\citep{bai2023qwenvl}, and are increasingly applied to aesthetic content as well, for example, to extract aesthetic attributes from images for multimodal recommendation~\citep{lee2026aesmmkg}. Despite these advances, current VLMs remain strong generalists rather than reliable specialists: they identify objects and produce fluent descriptions, but these strengths do not transfer to aesthetic visual understanding automatically~\citep{10.1145/3746027.3754961,wang2025aestest,10.1145/3664647.3680649}. Photographic critique demands a different kind of visual reasoning: the model must explain \textit{why} a framing choice helps or harms an image and \textit{how} lighting supports or undermines subject emphasis, and such judgments are further distorted by semantic-aesthetic bias, where visually appealing subjects receive disproportionately positive evaluations even when technical quality is poor~\citep{bulat2025beauty}.

Recent approaches fine-tune foundational VLMs for aesthetic critique, often leveraging newly curated datasets~\citep{10.1145/3664647.3680649, zhou2024uniaa, 10655670}, while others investigate reasoning strategies like chain-of-thought prompting and reinforcement learning to improve grounded critique~\citep{10.1145/3746027.3754961}. Architectures such as PhotoEye~\citep{11093854} combine multiple vision encoders with language models, showing that aesthetic assessment benefits from multiple viewpoints, and PhotoFramer~\citep{you2025photoframer} provides prescriptive composition guidance, generating re-framing instructions together with a well-composed example image. The Reddit Photo Critique Dataset (RPCD)~\citep{10.5555/3600270.3602745} captures community feedback from hobbyists to professional photographers, providing diverse critique styles particularly useful for LoRA-based fine-tuning. However, visual instruction tuning~\citep{liu2023visual} in this monolithic fashion can introduce catastrophic forgetting, degrading broader visual reasoning while improving the target task~\citep{ghosh24a, li2025dynamic}.

Score-based methods provide no actionable insight for photographers, and monolithic critique VLMs like PhotoEye rely on multi-encoder fusion with full-parameter fine-tuning on large curated corpora. SE-MoLoRA instead targets controllable, domain-specific critique generation through lightweight adapters on a single frozen backbone, preserving the base model's visual grounding. PhotoFramer is complementary: it prescribes how to re-frame a shot within the composition domain, whereas SE-MoLoRA critiques an existing photograph across multiple aesthetic domains through explicitly disentangled experts.

\subsection{Parameter-Efficient Adaptation and Mixture-of-LoRA Experts}

Low-Rank Adaptation (LoRA)~\citep{hu2022lora} has become the dominant parameter-efficient fine-tuning strategy, augmenting frozen models with trainable low-rank decomposition matrices that approximate full fine-tuning at negligible inference overhead once merged. The framework has since been extended in several directions: quantization-aware variants~\citep{10.5555/3666122.3666563} for memory efficiency, DoRA~\citep{10.5555/3692070.3693369} for expressivity through magnitude-direction decoupling, and LoRA+~\citep{10.5555/3692070.3692782} for improved convergence. An important consideration is rank selection: increasing rank improves adaptation capacity but raises overfitting risk~\citep{zhang2023adaptive}, motivating higher-rank shared components for cross-domain knowledge and lower-rank specialists for domain-specific residuals.

To route inputs to specialized adapters, we draw on Mixture-of-Experts (MoE) architectures, which enable conditional computation by gating a subset of experts per input~\citep{2017olln}. Recent adapter-centric implementations extend MoE from token-level feed-forward modules to modular adaptation layers, routing over LoRA blocks rather than dense experts~\citep{mixture}. While classical MoE routing operates at the token level, adapter-based routing occurs at the task or query level, making it suitable for domain-oriented applications where specialization is determined by overall input context. The DeepSeek MoE architecture~\citep{deepseekmoe} introduces shared experts alongside routed experts, providing always-active representations while routed experts specialize in domain-specific patterns; related designs combine shared and routed LoRA experts for multi-task and instruction tuning~\citep{yang2025adaptive, dou2024loramoe, tian2024hydralora, zhao2025mosld}. Modular MoE designs that pair explicitly domain-specific experts with a router have likewise been shown to preserve knowledge across languages and domains while mitigating catastrophic forgetting~\citep{almaamari2025mmoe}.

Unlike prior MoE-LoRA methods that primarily target multi-task or general instruction adaptation over cleanly separable tasks, SE-MoLoRA is designed for critique domains that are semantically overlapping. It adopts the shared-expert concept from DeepSeekMoE but implements it entirely through parameter-efficient LoRA adapters with query-level routing, enabling MoE-style specialization without MoE-scale compute. Its components thus build on existing shared-expert, routing, and orthogonality techniques; the contribution lies in adapting them to the decomposition of photographic critique into shared aesthetic language plus routed domain expertise.

\subsection{Residual Specialization and Expert Disentanglement}

Specialist adapters must learn distinct features to justify their modular cost. In modular adaptation, domain experts should learn incremental differences on top of shared knowledge rather than full mappings from scratch, following the principle that optimizing residual mappings is easier than optimizing unreferenced mappings~\citep{7780459}. Chaining multiple LoRA adapters through iterative residual updates has been shown to approximate full-parameter fine-tuning while maintaining computational efficiency~\citep{xia2024chain}. This hierarchy also addresses the stability-plasticity dilemma~\citep{1611835114}, where freezing the base model preserves prior capabilities while adapters provide integration of new domains.

However, if multiple specialists learn similar representations, modular capacity is wasted and every expert collapses into a shared expert. This risk is acute for photographic critique, where aesthetic attributes and semantic content are frequently entangled. Orthogonal regularization addresses this by penalizing cosine similarity between adapter weight matrices during training, inspired by concept-level disentanglement in VLMs~\citep{li_deal_2024} and DisLoRA~\citep{yifei-etal-2025-dislora}, which shows that enforcing orthogonal task-specific directions in LoRA updates prevents representational collapse. Closest to our sequential penalty, O-LoRA~\citep{wang2023olora} learns each new task in a subspace orthogonal to previously trained, frozen LoRA adapters for continual learning. A further challenge is spectral dominance, where a shared expert's activations may overtake specialist contributions during inference, addressable through linear weighting or model merging techniques like TIES~\citep{10.5555/3666122.3666432} and DARE~\citep{10.5555/3692070.3694452}.

SE-MoLoRA operationalizes residual specialization through a hierarchical pipeline: a frozen base model preserves general multimodal capabilities, a shared adapter captures cross-domain photographic knowledge and is merged into the base, and specialists then learn domain-specific residuals on top of it. Whereas DisLoRA enforces orthogonality across distinct tasks, our regularization separates specialists \emph{within} a single task whose domains inherently overlap, and asymmetric adapter weighting preserves specialist contributions against spectral dominance at inference.
\section{Methodology} \label{sec:methodology}

\subsection{LLM-Distilled RPCD Dataset} \label{sec:meth-dataset}

The first requirement is domain-labeled training data. We build on the Reddit Photo Critique Dataset (RPCD)~\citep{10.5555/3600270.3602745}, which pairs approximately 74,000 images with 220,000 free-form community critiques. Because raw comments are noisy and often span multiple aesthetic domains, we apply LLM-based distillation to decompose them into domain-specific training samples.

All comments associated with a given post are concatenated and processed with a constrained instruction prompt to decompose the critique into four expert domains. The seven original PCCD attributes~\citep{10.1145/3343031.3350970} are grouped by thematic similarity: \textit{general\_impression} and \textit{subject\_of\_photo} form the \textbf{shared} domain; \textit{composition} maps to the \textbf{composition} domain; \textit{color\_lighting} maps to the \textbf{lighting} domain; and \textit{focus}, \textit{depth\_of\_field}, and \textit{use\_of\_camera} form the \textbf{technical} domain. This process yields approximately 47,000 domain-labeled training samples. Distillation is performed with Claude Sonnet 4; raw comments are cleaned with the RPCD preprocessing pipeline~\citep{10.5555/3600270.3602745} before decomposition, and no additional filtering is applied. The prompt is deliberately \emph{extractive}, reorganizing existing comment content into domains rather than authoring new judgments, which bounds the distillation model's influence on the supervision signal.

\subsection{SE-MoLoRA Architecture} \label{sec:meth-architecture}

SE-MoLoRA decomposes a single monolithic adapter into a permanently active shared expert and a set of conditionally activated specialist experts. The architecture is built on top of Qwen3-VL-8B~\citep{bai2023qwenvl}, which serves as the frozen base model.

\begin{figure}[t]
    \centering
    \includegraphics[width=\columnwidth]{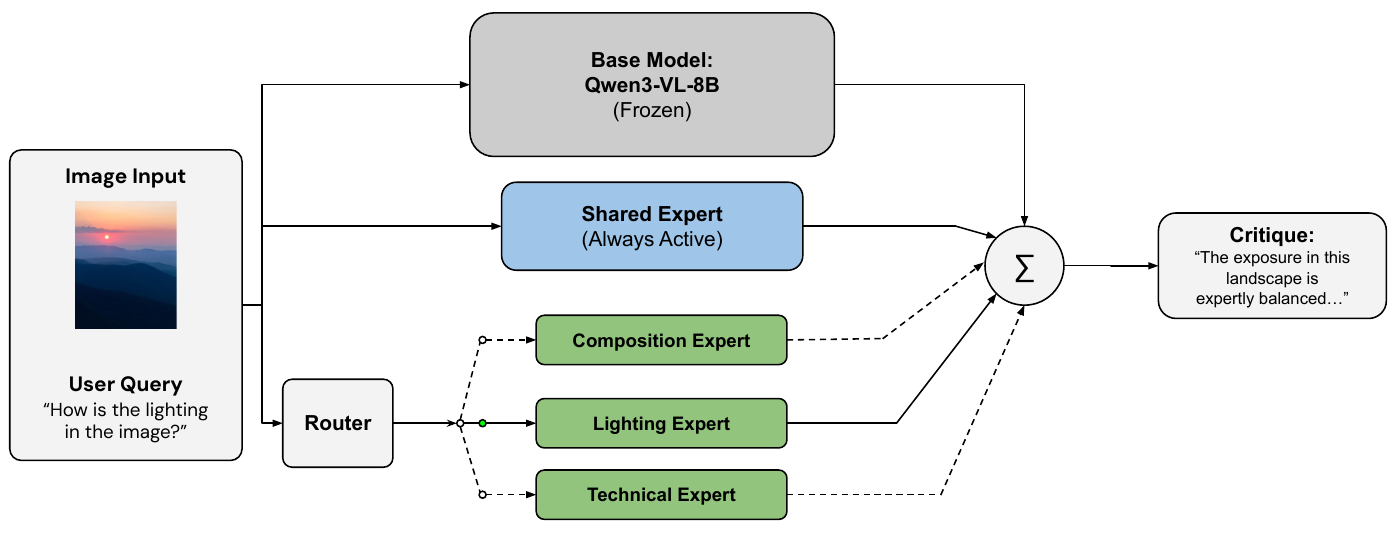}
    \caption{The SE-MoLoRA architecture. A shared adapter handles general photographic knowledge, while specialized adapters for composition, lighting, and technical aspects are conditionally activated based on the router's decision.}
    \Description{Block diagram of SE-MoLoRA: a frozen Qwen3-VL base model with an always-active shared LoRA adapter, three specialist LoRA adapters for composition, lighting and technical quality, and a router that selects one specialist from the image and query.}
    \label{fig:se-molora-architecture}
\end{figure}

As shown in Figure~\ref{fig:se-molora-architecture}, the system consists of three layers: (1) the frozen base model $W_{\text{base}}$, (2) a shared LoRA adapter $\Delta W_s$ (rank-64) that is always active and captures domain-agnostic photographic vocabulary, and (3) three specialist LoRA adapters (rank-32 each) for composition, lighting, and technical analysis, of which at most one is activated per query. The rank asymmetry mirrors the knowledge hierarchy: the shared adapter must capture broad cross-domain photographic vocabulary and therefore receives higher capacity, whereas the specialists learn only residuals on top of the merged shared expert, for which lower rank suffices (validated empirically in Section~\ref{sec:res-quality}); larger ranks would also raise overfitting risk~\citep{zhang2023adaptive}. The total weight during inference is: 

\begin{equation}\label{eq:weight}
W_{\text{out}} = W_{\text{base}} + \lambda_s \Delta W_s + \lambda_e \Delta W_{e^*}, \qquad e^* = \operatorname{Route}(q),
\end{equation}
where $\lambda_s$ and $\lambda_e$ control the relative influence of the shared and specialist adapters, and $e^*$ is the specialist selected by the router for query $q$.

The \textbf{shared expert} is trained on the general\_impression and subject\_of\_photo domains, aligning the model to professional photography terminology without introducing aesthetic bias. It also serves as a fallback for ambiguous queries that cannot be routed to a specialist. The \textbf{specialist experts} each focus on distinct aesthetic attributes: the composition expert targets framing, leading lines, and visual balance; the lighting expert targets exposure, white balance, and color grading; and the technical expert targets focus accuracy, depth of field, noise, and lens artifacts.

\subsection{Training Strategy} \label{sec:meth-training}

\textbf{Three-stage pipeline.} Training all adapters jointly risks the same domain entanglement the architecture aims to eliminate. We therefore adopt a three-stage residual learning pipeline. In \textit{Stage1}, the shared expert adapter is trained on approximately 15K samples using standard supervised fine-tuning with causal language modeling. In \textit{Stage2}, the trained shared adapter is merged into the base model weights, producing an enhanced base model $W_{\text{new}} = W_{\text{base}} + \Delta W_s$. This merging step ensures that specialist gradients in Stage3 flow only through their own low-rank matrices rather than through the shared adapter, preventing vanishing gradients and interference issues. In \textit{Stage3}, each specialist adapter is trained independently on 5K domain-specific samples on top of $W_{\text{new}}$. Specialists never observe each other's gradients, minimizing cross-domain interference. At inference, the shared adapter is not used in merged form: it is applied as a separate adapter alongside the selected specialist following Equation~\ref{eq:weight}. Each specialist, trained on top of the fully weighted shared adapter, therefore runs with the shared component down-weighted to $\lambda_s$.

\textbf{Orthogonal regularization.} Despite domain-separated training data, specialist adapters can still learn overlapping representations. To enforce disentanglement, we add an orthogonal regularization term in the standard loss function (Equation~\ref{eq:total_loss}) that penalizes cosine similarity between specialist weight updates (Equation~\ref{eq:ortho_loss}): 
\begin{equation}\label{eq:total_loss} 
\mathcal{L}_{\text{total}} = \mathcal{L}_{\text{SFT}} + \lambda_{\text{ortho}} \cdot \mathcal{L}_{\text{ortho}} 
\end{equation} 

\begin{equation}\label{eq:ortho_loss} 
\mathcal{L}_{\text{ortho}} = \sum_{l \in L} \frac{\bigl|\operatorname{vec}(\Delta W_{\text{curr}}^{(l)}) \cdot \operatorname{vec}(\Delta W_{\text{ref}}^{(l)})\bigr|}{\|\Delta W_{\text{curr}}^{(l)}\|_F \, \|\Delta W_{\text{ref}}^{(l)}\|_F}
\end{equation} 

where $\Delta W_{\text{curr}}$ is the adapter being trained and $\Delta W_{\text{ref}}$ are the weights of previously trained specialists. Specialists are trained sequentially: the composition expert is trained first, then the lighting expert uses the composition weights as reference, and the technical expert uses both.

\textbf{Spectral dominance mitigation.} Due to the rank asymmetry between the shared (rank-64) and specialist (rank-32) adapters, the shared adapter's larger singular values can dominate the final output, drowning out specialist contributions. We address this with asymmetric weighting: $\lambda_s = 0.5$ and $\lambda_e = 1.5$ in Equation~\ref{eq:weight}, which rebalances the relative influence during inference. These values were chosen heuristically by inspecting generated critiques rather than tuned on a held-out split.

\subsection{Routing Mechanism} \label{sec:meth-routing}

At inference time, the system must decide which specialist to activate based on the user's query. Unlike token-level MoE routing~\citep{2017olln}, our routing operates at the query level to determine overall user intent. SE-MoLoRA employs a multimodal router for this decision; to isolate the contribution of visual cues, we additionally train a text-only router as a baseline and compare both:

\textbf{Text-only router.} A fine-tuned DistilBERT model (67M parameters) performs four-class classification over the expert domains. The training data combines combinatorially generated prompt-label pairs from domain-specific vocabulary with naturalistic queries from the PCCD dataset~\citep{10.1145/3343031.3350970} to balance keyword coverage with linguistic variability. This router is fast and lightweight but cannot leverage visual cues when user prompts are vague.

\textbf{Multimodal router.} To handle ambiguous queries where the image itself contains diagnostic cues (e.g., ``What do you think?'' paired with an overexposed image), we introduce a CLIP ViT-B/32-based multimodal classifier. Frozen CLIP encoders extract image and text embeddings in parallel, which are normalized, concatenated, and passed through a trainable MLP classification head (260K trainable parameters). Training labels are derived from PCCD attribute scores: images with disproportionately low scores in a specific attribute are labeled as requiring that domain's specialist.
\section{Experiments}
\label{sec:result}

\subsection{Implementation Details}

All experiments use a single NVIDIA A100 (40GB) with bfloat16 precision, Unsloth's memory-optimized training, and 8-bit AdamW. The shared expert (rank-64) trains on 15K samples for 5 epochs; each specialist (rank-32) trains on 5K domain-specific samples for 3 epochs. The base model is Qwen3-VL-8B-Instruct~\citep{bai2023qwenvl}. Each domain's distilled data is split 80/10/10 into training, validation, and test sets (seed 42); the checkpoint with the lowest validation loss is kept. Each adapter is trained once (seed 3407) and decoded greedily, so significance tests are paired over test samples rather than seeds. The monolithic baseline (rank 64) is trained for 5 epochs on the concatenated critiques of all domains under a generic critique prompt and, like the experts, receives the domain-specific instruction at test time. Unless noted otherwise, results use oracle routing, i.e., each test sample is answered by the expert of its labeled domain; router accuracy is analyzed separately in Section~\ref{sec:res-router}.

\subsection{Critique Quality and Domain Specificity} \label{sec:res-quality}

We first evaluate whether modular adaptation improves critique alignment over monolithic training. Table~\ref{tab:automated_metrics} compares SE-MoLoRA against the zero-shot base model and a monolithic LoRA baseline trained on the same combined data. The shared expert achieves a BERTScore~F1 of $0.4215$, roughly ten times the base model ($0.0379$) and nearly double the monolithic baseline ($0.2317$). All three specialists likewise outperform both baselines across every metric, with composition ($0.2999$) and lighting ($0.2990$) showing the largest gains over the monolithic adapter. The base model produces negative BERTScore~F1 on all specialist domains, indicating that its verbose outputs are less aligned with reference critiques than even the rescaled baseline. All improvements are statistically significant ($p < 0.001$, paired t-test).

\begin{table*}[t]
\caption{Automated metric evaluation comparing the zero-shot Base model, a Monolithic LoRA baseline, and our routed SE-MoLoRA architecture. Best results are highlighted in bold. A paired Student's t-test confirmed statistical significance ($p < 0.001$).}
\label{tab:automated_metrics}
\centering\footnotesize
\begin{tabular}{llccc}
\toprule
\textbf{Expert} & \textbf{Architecture} & \textbf{BERTScore F1} ($\uparrow$) & \textbf{SBERT (Cos Sim)} ($\uparrow$) & \textbf{ROUGE-L F1} ($\uparrow$)\\
\midrule
\multirow{3}{*}{\textbf{Shared}} 
 & Base Model & 0.0379 $\pm$ 0.0751 & 0.6105 $\pm$ 0.1140 & 0.0921 $\pm$ 0.0309 \\
 & Monolithic & 0.2317 $\pm$ 0.0831 & 0.6337 $\pm$ 0.1013 & 0.1787 $\pm$ 0.0629 \\
 & \textbf{SE-MoLoRA} & \textbf{0.4215 $\pm$ 0.1000} & \textbf{0.6851 $\pm$ 0.1061} & \textbf{0.3339 $\pm$ 0.1018} \\
\midrule
\multirow{3}{*}{\textbf{Composition}} 
 & Base Model & -0.0211 $\pm$ 0.0478 & 0.5277 $\pm$ 0.1232 & 0.0810 $\pm$ 0.0270 \\
 & Monolithic & 0.2348 $\pm$ 0.0773 & 0.5362 $\pm$ 0.1265 & 0.1845 $\pm$ 0.0530 \\
 & \textbf{SE-MoLoRA} & \textbf{0.2999 $\pm$ 0.0745} & \textbf{0.5851 $\pm$ 0.1336} & \textbf{0.2145 $\pm$ 0.0573} \\
\midrule
\multirow{3}{*}{\textbf{Lighting}} 
 & Base Model & -0.0741 $\pm$ 0.0456 & 0.5155 $\pm$ 0.1230 & 0.0414 $\pm$ 0.0168 \\
 & Monolithic & 0.1741 $\pm$ 0.0938 & 0.5253 $\pm$ 0.1145 & 0.1427 $\pm$ 0.0520 \\
 & \textbf{SE-MoLoRA} & \textbf{0.2990 $\pm$ 0.1060} & \textbf{0.5410 $\pm$ 0.1428} & \textbf{0.2014 $\pm$ 0.0905} \\
\midrule
\multirow{3}{*}{\textbf{Technical}} 
 & Base Model & -0.0488 $\pm$ 0.0496 & 0.4467 $\pm$ 0.1248 & 0.0677 $\pm$ 0.0238 \\
 & Monolithic & 0.2132 $\pm$ 0.0823 & 0.3919 $\pm$ 0.1367 & 0.1355 $\pm$ 0.0566 \\
 & \textbf{SE-MoLoRA} & \textbf{0.2461 $\pm$ 0.0796} & \textbf{0.4992 $\pm$ 0.1345} & \textbf{0.1727 $\pm$ 0.0571} \\
\bottomrule
\end{tabular}
\end{table*}

\begin{figure}[t]
    \centering
    \includegraphics[width=\columnwidth]{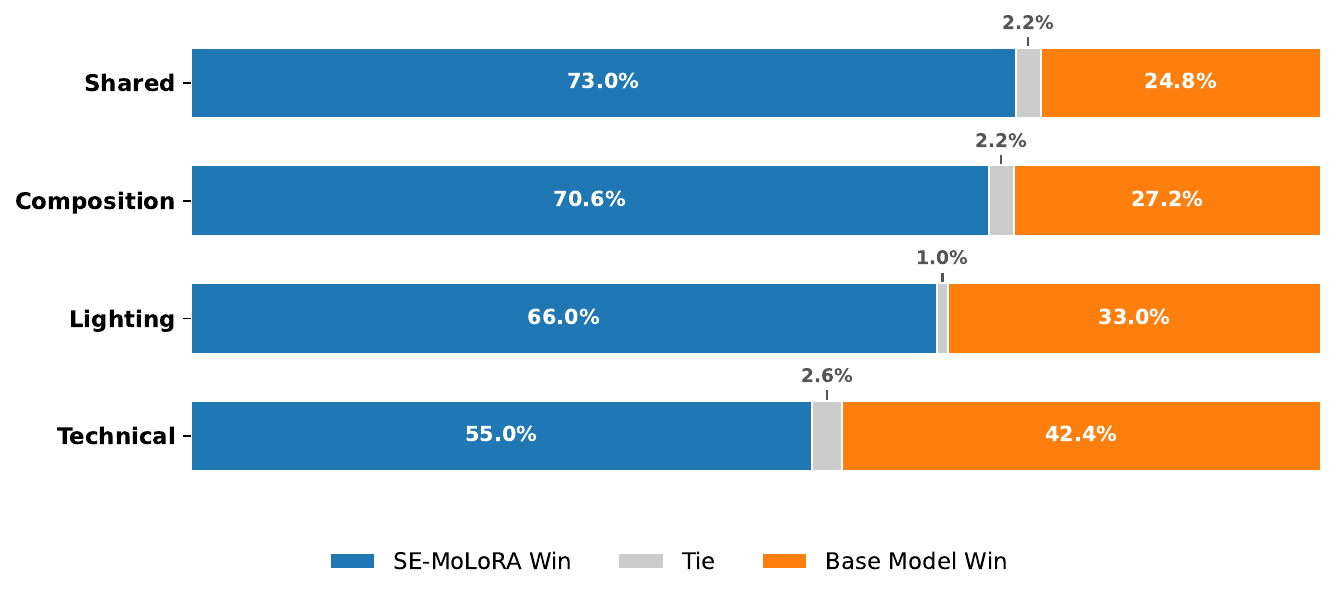}\\[1mm]
    \includegraphics[width=\columnwidth]{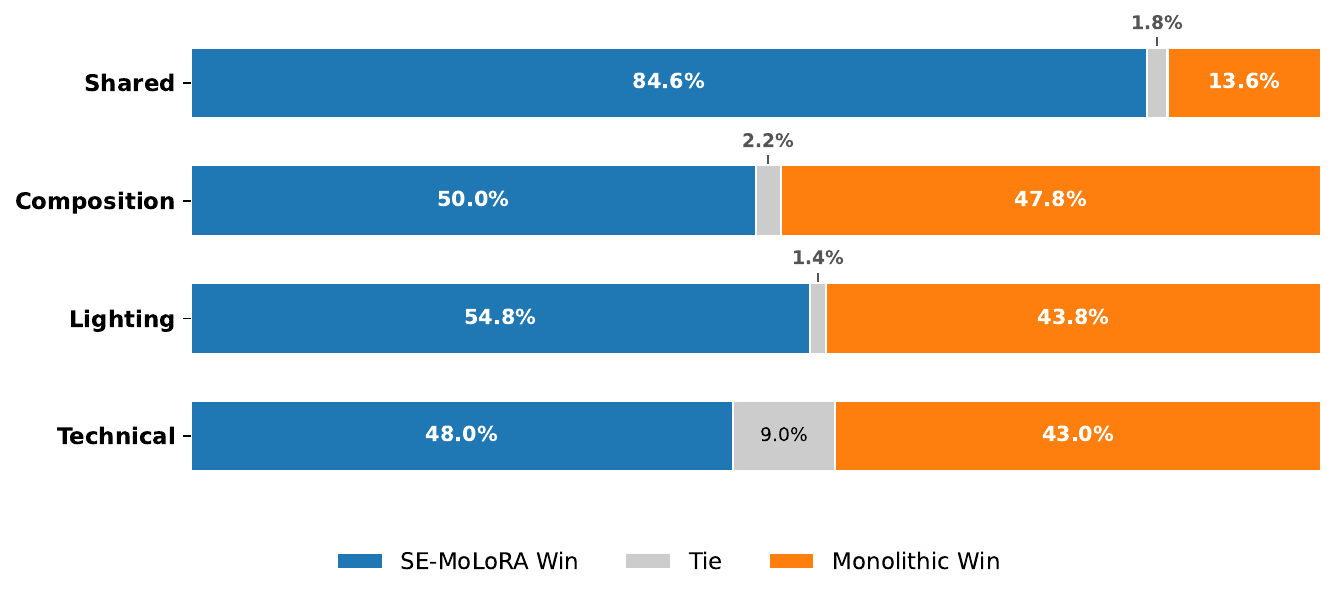}
    \caption{LLM-as-a-Judge Evaluation Win Rates of the SE-MoLoRA architecture against the zero-shot base model (Top) and the monolithic adapter (Bottom).}
    \Description{Two horizontal bar charts of win, tie, and loss rates for the shared, composition, lighting, and technical experts against the base model and against the monolithic adapter.}
    \label{fig:llm_judge_results}
\end{figure}

To complement reference-based evaluation, we conduct blind pairwise comparisons using GPT-4o as an LLM-as-judge (Figure~\ref{fig:llm_judge_results}). Comparisons are position-randomized with an anti-verbosity judge prompt, following protocols shown to correlate strongly with human preference~\citep{NEURIPS2023_91f18a12} and to be robust when recent, large models serve as judges~\citep{degiovanni2026judge}. The reference critiques are Claude-restructured versions of human community comments, so both reference-based metrics and the judge may partly reward matching the distilled style; the judge belongs to a different model family than the distillation model, which reduces self-preference toward that style. Against the base model, all four experts achieve decisive win rates. Against the monolithic baseline, the shared expert wins $84.6\%$ of comparisons, demonstrating that monolithic fine-tuning over-specializes and cannot suppress domain-specific knowledge when tasked with general critique. To separate the effect of decomposition from capacity, we additionally evaluate monolithic adapters at ranks 32 and 96 (Table~\ref{tab:rank_sweep}). The composition and lighting specialists clearly beat the equal-capacity rank-32 monolithic adapter ($59.2\%$ and $63.6\%$), and remain at parity with the rank-64 and rank-96 adapters (mean win rates of $50.9\%$ and $52.5\%$), the latter matching SE-MoLoRA's active rank of $64{+}32$; with 500 pairs, each win rate carries a 95\% confidence interval of at most $\pm4.4$ points. The shared expert wins decisively at every rank. The technical expert remains weakest, consistent with the lower share of technical content in the distilled data. Raising the monolithic rank from 64 to 96 yields no consistent gain, so additional capacity alone does not close the gap to the shared expert. We also note that Table~\ref{tab:automated_metrics} and Figure~\ref{fig:llm_judge_results} probe different constructs: automated metrics measure alignment with reference critiques and are sensitive to verbosity suppression and domain vocabulary, hence the large margins, whereas a blind pairwise judge may credit an on-topic monolithic answer even when it is less domain-precise, narrowing the margin.

\begin{table}[t]
\caption{Win rates (\%) of SE-MoLoRA experts against monolithic LoRA baselines at ranks 32, 64, and 96 (500 pairs each).}
\label{tab:rank_sweep}
\centering
\begin{tabular}{lccc}
\toprule
\textbf{Expert win \% vs.} & \textbf{r-32} & \textbf{r-64} & \textbf{r-96} \\
\midrule
Shared      & 74.6 & 84.6 & 82.8 \\
Composition & 59.2 & 50.0 & 50.5 \\
Lighting    & 63.6 & 54.8 & 48.8 \\
Technical   & 46.3 & 48.0 & 58.1 \\
\bottomrule
\end{tabular}
\end{table}

\begin{figure*}[t]
    \centering
    \includegraphics[width=0.8\textwidth]{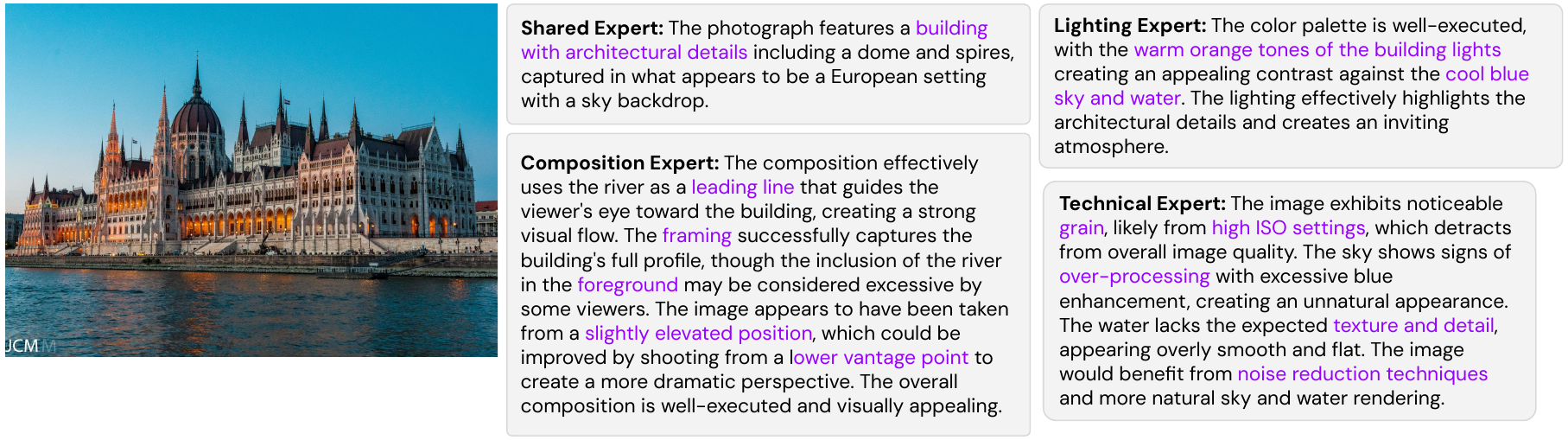}
    \caption{Qualitative comparison between responses from each expert for the same image (Sample taken from RPCD~\citep{10.5555/3600270.3602745}). Domain-specific vocabulary for each expert is highlighted in purple.}
    \Description{One photograph next to four critiques produced by the shared, composition, lighting, and technical experts, with domain-specific vocabulary highlighted.}
    \label{fig:all-experts}
\end{figure*}

While these aggregate comparisons confirm quality gains, they do not reveal whether specialists produce qualitatively \emph{distinct} outputs. To test this, we perform a switch test: identical images are presented to the system with each expert activated in turn. As shown in Figure~\ref{fig:all-experts}, each expert critiques with visibly distinct domain-specific vocabulary, confirming functional specialization.

\subsection{Expert Disentanglement} \label{sec:res-disentangle}

Improved critique metrics do not guarantee that specialists encode distinct knowledge. To verify structural disentanglement, we analyze the learned weight updates $\Delta W = BA$: for each adapted module we take the principal left singular vector of $\Delta W$, compute the absolute cosine similarity between two adapters' vectors, and average over all modules. Table~\ref{tab:entanglement_ablation} presents entanglement values across four configurations of the orthogonal penalty $\lambda_{\text{ortho}}$.

\begin{table}[t]
\caption{SVD-based cosine similarity between expert adapters. Lower values indicate greater disentanglement. SE refers to Shared Expert.}
\label{tab:entanglement_ablation}
\centering\footnotesize
\begin{tabular}{lccccc}
\toprule
& \textbf{No SE} & \multicolumn{4}{c}{\textbf{SE present during training}} \\
\cmidrule(lr){3-6} \textbf{Expert Pair} & \textbf{during training} & \textbf{$\lambda=0.0$} & \textbf{$\lambda=0.1$} & \textbf{$\lambda=0.5$} & \textbf{$\lambda=1.0$} \\
\midrule Comp. vs Light. & 0.6468 & 0.5395 & 0.4365 & \textbf{0.4358} & 0.4392 \\
Comp. vs Tech. & 0.6241 & 0.5169 & 0.4229 & \textbf{0.4193} & 0.4233 \\
Light. vs Tech. & 0.6380 & 0.5251 & 0.4238 & 0.4230 & \textbf{0.4229} \\ 
\midrule 
Shared vs Comp. & 0.4335 & 0.3512 & 0.3477 & 0.3477 & 0.3477 \\
Shared vs Light. & 0.4314 & 0.3426 & 0.3419 & 0.3420 & \textbf{0.3408} \\ 
Shared vs Tech. & 0.4223 & 0.3461 & 0.3418 & \textbf{0.3392} & 0.3422 \\
\bottomrule
\end{tabular}
\end{table}

Without orthogonal regularization, specialist pairs exhibit high entanglement ($\approx 0.52$). Applying $\lambda_{\text{ortho}}=0.1$ reduces this to $\approx 0.43$, but increasing the penalty beyond this yields diminishing returns, establishing an entanglement floor. This floor reflects a fundamental interdependence of photographic critique domains: low-level visual features such as edge detection are shared across composition, focus, and lighting analysis.

The shared expert acts as an entanglement sink, absorbing common features as evidenced by consistently lower shared-specialist similarity ($\approx 0.34$) compared to specialist-specialist pairs ($\approx 0.43$). Removing the shared expert entirely (first column) increases all pairwise similarities substantially, confirming its role in promoting disentanglement.

\subsection{Routing Accuracy and Benchmark Positioning} \label{sec:res-router}

On the held-out test set, the text-only router achieves higher classification accuracy than the multimodal router ($93.42\%$ vs.\ $86.95\%$), as routing labels are heavily correlated with textual vocabulary.

However, on downstream aesthetic benchmarks with naturalistic queries (Table~\ref{tab:aesthetic_benchmarks}), the multimodal router consistently outperforms the text-only router. On Q-Bench, the multimodal router achieves $57.65\%$ vs.\ $55.51\%$; on PhotoBench, $53.97\%$ vs.\ $50.66\%$. This reversal occurs because benchmark prompts are generally vague, and the text-only router lacks visual grounding to detect latent image flaws, defaulting to the shared expert. The multimodal router detects visual outliers from the image, routing generic queries to suitable specialists at the cost of higher inference latency ($3.60 \pm 0.45$ vs.\ $2.17 \pm 0.18$ ms/sample). SE-MoLoRA therefore adopts the multimodal router by default, and all SE-MoLoRA benchmark results below report this configuration; the text-only router remains a lower-latency alternative when queries are explicit.

A per-class error analysis also exposes the dominant failure mode of each router. The text-only router misroutes generically phrased technical queries to the shared expert (46 cases) and, on vague prompts, defaults to the shared expert almost exclusively, which explains its weaker benchmark scores. The multimodal router shows the inverse failure: genuinely general queries are over-routed to the technical expert when the image contains visible artifacts (127 shared$\to$technical confusions). The same visual flaw detection that helps it on benchmarks misfires when the user's intent is explicitly general.

\begin{table}[t]
\caption{Router comparison on zero-shot aesthetic benchmarks.}
\label{tab:aesthetic_benchmarks}
\centering
\begin{tabular}{lcc}
\toprule
\textbf{Benchmark} & \textbf{Text Router} & \textbf{Multimodal Router} \\
\midrule 
Q-Bench & 55.51\% & \textbf{57.65\%} \\ 
PhotoBench & 50.66\% & \textbf{53.97\%} \\
\bottomrule
\end{tabular}
\end{table}

\subsubsection{Positioning on zero-shot benchmarks}
Both benchmarks above also allow positioning SE-MoLoRA against published aesthetic-assessment systems, with an important caveat stated up front: Q-Bench and PhotoBench are zero-shot multiple-choice QA tasks, a different operating point from the open-ended, domain-controllable critique generation this system targets. Generic benchmark queries rarely trigger the intended specialist, so the system falls back to the shared expert. Each answer option is scored by its perplexity under the model and the lowest-perplexity option is selected, so no output parsing is involved; this likelihood-based protocol differs from the generation-based protocols of prior reports, so absolute scores are not directly comparable.

SE-MoLoRA improves substantially over its own zero-shot base model ($+6.9$ points on Q-Bench and $+22.3$ on PhotoBench), indicating that the adapters transfer useful aesthetic knowledge even under this unfavorable query format, and on PhotoBench it surpasses two published aesthetic systems, UNIAA~\citep{zhou2024uniaa} ($+17.1$ points) and Q-Instruct~\citep{10655670} ($+10.1$ points), as reported by \citet{11093854} on older backbones, so this comparison is indicative only. It underperforms benchmark-optimized systems such as PhotoEye~\citep{11093854}. The remaining gap reflects a deliberate resource trade-off: PhotoEye employs continuous multi-view vision fusion from diverse foundation models (CLIP, DINOv2, SAM, CoDETR) with full-parameter fine-tuning on 450K curated images, whereas SE-MoLoRA freezes the base encoder and adapts only through lightweight LoRA modules trained on 47K distilled samples.

\subsection{Ablation: Impact of the Shared Expert} \label{sec:res-ablation}

\begin{figure}[t]
    \centering
    \includegraphics[width=\columnwidth]{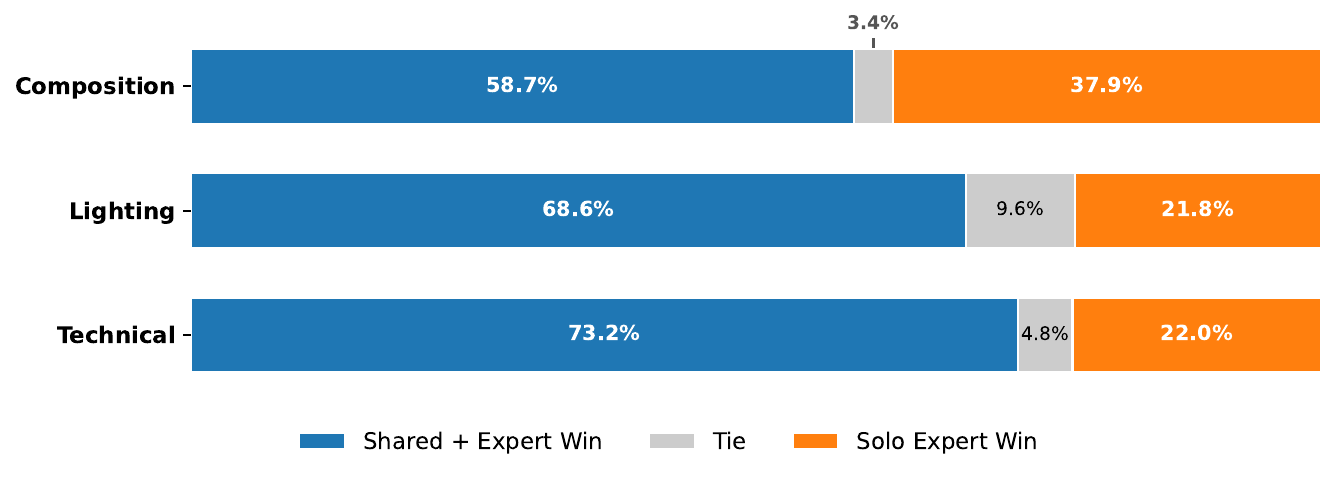}
    \caption{Ablation results showing the impact of the shared expert on win rates with LLM-as-judge.}
    \Description{Horizontal bar chart of LLM-judge win rates for specialists trained with versus without the shared expert.}
    \label{fig:without_shared_expert}
\end{figure}

To isolate the shared expert's contribution, we compare the full SE-MoLoRA pipeline against a variant where specialist adapters are trained directly on the base model without the shared expert. Using GPT-4o as an LLM-as-judge (Figure~\ref{fig:without_shared_expert}), specialists trained with the shared expert produce consistently higher win rates across all domains, generating critiques with proper visual grounding that bridge general visual context with domain-specific analysis. The disentanglement data in Table~\ref{tab:entanglement_ablation} complements this finding: without the shared expert, specialist pairs are substantially more entangled ($\approx 0.64$ vs.\ $\approx 0.43$), confirming that the shared expert absorbs cross-domain features and prevents specialists from wasting capacity on redundant representations. The shared expert particularly benefits the technical domain, which otherwise shows the lowest gains across all metrics.

\section{Conclusion}
\label{sec:conclusion}


Our experiments demonstrate that decomposing photographic critique into modular LoRA adapters yields measurably stronger reference alignment than monolithic fine-tuning: specialists outperform an equal-capacity monolithic adapter and match monolithic adapters of up to the system's full active rank, while the shared expert is preferred at every rank. The shared expert is essential: it captures cross-domain features that would otherwise become entangled within the specialists, reducing pairwise cosine similarity from $\approx 0.64$ to $\approx 0.43$ together with orthogonal regularization, while improving visual grounding across all domains. For real-world deployment, the multimodal router enables more robust specialist selection for vague queries, although it incurs higher latency than the text-only alternative.


\textbf{Limitations and future work.} The current system activates only one specialist per query, limiting the number of active adapters to two and preserving the interpretability of domain attribution. However, the formulation in Equation~\ref{eq:weight} naturally supports weighted combinations of multiple specialists without requiring architectural changes. Replacing the hard router with soft classification over multiple experts therefore represents a natural next step, while multi-specialist activation is left for future work.


All evaluations rely on automated metrics and LLM-based judgments rather than human user studies. Evaluations involving professional photographers would help establish whether disentangled critiques are perceived as more actionable. Moreover, the reference critiques are Claude-restructured versions of human comments, so reference-based metrics may favor models trained on the same distilled format. The comparison is also limited to zero-shot and monolithic baselines; MoE-LoRA methods, independently trained per-domain adapters, and published critique models evaluated on the generation task itself are left for future work, as is end-to-end evaluation with the learned router. The training data also carries sourcing bias. RPCD critiques are English-only, reflect community-specific photographic conventions, and rarely discuss concrete camera settings, which likely contributes to the comparatively smaller gains achieved by the technical expert. Furthermore, the architecture is evaluated using a single base model (Qwen3-VL-8B) and three aesthetic domains, while the inference weighting coefficients ($\lambda_s$, $\lambda_e$) were set heuristically for this backbone, the shared adapter is down-weighted at inference relative to training ($\lambda_s = 0.5$ vs.\ $1$) without an ablation of this mismatch, and each configuration was trained with a single seed. Generalization to additional domains and base models therefore remains unexplored. Finally, the SE-MoLoRA framework is not limited to photographic critique and may be applicable to other domains in which specialist knowledge exhibits varying degrees of interdependence.

\begin{acks}
This research has been supported by the Business Finland-funded AI-Native Software Engineering (ANSE) project under the grant diary number 1822/31/2025.
\end{acks}
\balance
\printbibliography

\ifappendix
\appendix

\section{Data Distillation}
\label{appen:data_distillation}

The following prompts were used to distill raw comments from Reddit posts into structured output for each of the four experts. Figure~\ref{fig:distill_process} illustrates the system prompt, and Figure~\ref{fig:json_prompt} presents the user prompt used for this process.

\begin{figure}[tbp]
    \centering
\begin{tcolorbox}[
    colback=gray!5,
    colframe=gray!30,
    width=\linewidth,
    boxrule=0.5pt,
    arc=0mm
]
\begin{lstlisting}[basicstyle=\scriptsize\ttfamily, breaklines=true, frame=none]
You are a Lead Photography Editor training specialized AI experts. 
Your goal is to synthesize noisy social media comments into distinct, highly specialized critiques.

Rules:
1. Synthesize: Combine related comments. Ignore low-effort/off-topic chatter.
2. Nulls: If a category is not mentioned, return null.
3. STRICT PERSONA SEPARATION: You must write each critique using a completely different vocabulary, tone, and sentence structure. 
  - Shared Critique: Write like a police report. Dry, factual, purely descriptive. No adjectives. 
  - Composition: Write like an eccentric art professor. Focus on geometry, flow, and visual weight. Use dynamic, structural language. 
  - Lighting/Color: Write like a cinematic colorist. Focus on mood, photons, shadows, and color theory. Use emotive, light-based language. 
  - Technical: Write like a sterile camera manual. Focus on optics, sensors, and physics. Use harsh, technical terminology. 
\end{lstlisting}
\end{tcolorbox}
    \caption{System prompt used for data distillation from RPCD dataset.}
    \Description{Verbatim system prompt used to distill Reddit comments into four persona-separated critiques.}
    \label{fig:distill_process}
\end{figure}

\begin{figure}[tbp]
    \centering
\begin{tcolorbox}[
    colback=gray!5,
    colframe=gray!30,
    width=\linewidth,
    boxrule=0.5pt,
    arc=0mm
]
\begin{lstlisting}[basicstyle=\scriptsize\ttfamily, breaklines=true, frame=none]
Analyze the user comments and output a JSON object. Adhere strictly to the requested personas for each field.

User Comments:
- {comment_text}

Required JSON Structure:
{
  "shared_critique": "[Factual, dry description of subject and setting]",
  "composition_critique": "[Artistic, geometric critique of framing and perspective]",
  "lighting_color_critique": "[Cinematic analysis of exposure and color grading]",
  "technical_critique": "[Sterile, optical breakdown of focus, noise, and artifacts]"
}

Return ONLY the JSON object above. Do not include any other text or formatting.
\end{lstlisting}
\end{tcolorbox}
    \caption{User prompt for generating the structured JSON output.}
    \Description{Verbatim user prompt requesting a JSON object with shared, composition, lighting, and technical critiques.}
    \label{fig:json_prompt}
\end{figure}


\section{Training Prompts}
\label{appen:training_prompt}

The following prompts were used to train the respective experts. The specific configurations for the Shared, Composition, Lighting, and Technical experts are detailed in Figure~\ref{fig:training_prompt}.

\begin{figure}[tbp]
    \centering
\begin{tcolorbox}[
    colback=gray!5,
    colframe=gray!30,
    width=\linewidth,
    boxrule=0.5pt,
    arc=0mm
]
\begin{lstlisting}[basicstyle=\scriptsize\ttfamily, breaklines=true, frame=none]
SHARED_EXPERT_PROMPT = """
Provide a detailed visual description of the subject, setting, and content. Do not offer aesthetic judgment or critique.
"""

COMPOSITION_EXPERT_PROMPT = """
Act as a strict structural composition judge. Evaluate the visual geometry of this image. Focus exclusively on framing, rule of thirds, leading lines, and perspective. Do not mention color or light.
"""

LIGHTING_EXPERT_PROMPT = """
You are a master gaffer and colorist. Break down the illumination in this scene. Detail the exposure, contrast, color grading, and white balance. Ignore what the subject is doing.
"""

TECHNICAL_EXPERT_PROMPT = """
Perform a camera technical analysis. I need a review of the focal sharpness, ISO noise levels, likely lens distortion, and shutter speed artifacts. Limit your response to hardware and physics.
"""
\end{lstlisting}
\end{tcolorbox}
    \caption{Training prompt for each expert.}
    \Description{The four instruction prompts used to train the shared, composition, lighting, and technical experts.}
    \label{fig:training_prompt}
\end{figure}


\section{LLM-as-judge}
\label{appen:llm_judge}
\label{appen:llm_as_judge}

The qualitative evaluation was conducted using GPT-4o as an automated judge. Depending on the specific configuration being tested, the system prompt was adjusted to focus on different performance metrics: Base vs. Expert (Figure~\ref{fig:judge_base_vs_expert}), Monolithic vs. Expert (Figure~\ref{fig:judge_monolithic_vs_expert}), and Solo vs. Shared Expert (Figure~\ref{fig:judge_solo_vs_shared}). The final evaluation wrapper used to present candidates to the judge is shown in Figure~\ref{fig:judge_wrapper}.

\begin{figure}[tbp]
    \centering
\begin{tcolorbox}[
    colback=gray!5,
    colframe=gray!30,
    width=\linewidth,
    boxrule=0.5pt,
    arc=0mm
]
\begin{lstlisting}[basicstyle=\scriptsize\ttfamily, breaklines=true, frame=none]
# BASE VS EXPERT SYSTEM PROMPT
You are an elite Photography Editor evaluating AI-generated image critiques. 
Your job is to compare two candidate critiques against a "Gold Standard" Ground Truth.

CRITICAL RULES:
1. THE VERBOSITY PENALTY: The Gold Standard represents the ideal concise, synthesized length. Actively penalize candidates that generate unnecessarily long, chatty walls of text or conversational filler (e.g., "Based on the image...").
2. CONTENT ACCURACY: The winning candidate must accurately capture the specific visual critiques present in the Gold Standard.
3. STRICT PERSONA: Penalize candidates that start talking about unrelated concepts.

You must output a strictly formatted JSON object with your step-by-step reasoning and your final decision ("A", "B", or "Tie").
\end{lstlisting}
\end{tcolorbox}
    \caption{LLM-as-judge system prompt for Base vs. Expert comparison.}
    \Description{Verbatim system prompt given to the LLM judge.}
    \label{fig:judge_base_vs_expert}
\end{figure}

\begin{figure}[tbp]
    \centering
\begin{tcolorbox}[
    colback=gray!5,
    colframe=gray!30,
    width=\linewidth,
    boxrule=0.5pt,
    arc=0mm
]
\begin{lstlisting}[basicstyle=\scriptsize\ttfamily, breaklines=true, frame=none]
# MONOLITHIC VS EXPERT SYSTEM PROMPT
You are an elite Photography Editor evaluating AI-generated image critiques. 
Your job is to compare two candidate critiques against a "Gold Standard" Ground Truth.

CRITICAL RULES:
1. CONCEPT BLEED PENALTY: We are testing for strict task adherence. Actively penalize candidates that bleed into unrelated topics (e.g., discussing composition when the Gold Standard only discusses lighting).
2. DETAIL FIDELITY: Reward the candidate that captures the nuanced technical details present in the Gold Standard without drifting off-topic.
3. CONCISENESS: Both should be concise. Do not reward verbosity.

You must output a strictly formatted JSON object with your step-by-step reasoning and your final decision ("A", "B", or "Tie").
\end{lstlisting}
\end{tcolorbox}
    \caption{LLM-as-judge system prompt for Monolithic vs. Expert comparison.}
    \Description{Verbatim system prompt given to the LLM judge.}
    \label{fig:judge_monolithic_vs_expert}
\end{figure}

\begin{figure}[tbp]
    \centering
\begin{tcolorbox}[
    colback=gray!5,
    colframe=gray!30,
    width=\linewidth,
    boxrule=0.5pt,
    arc=0mm
]
\begin{lstlisting}[basicstyle=\scriptsize\ttfamily, breaklines=true, frame=none]
# SOLO VS SHARED EXPERT SYSTEM PROMPT
You are an elite Photography Editor evaluating AI-generated image critiques. 
Your job is to compare two candidate critiques against a "Gold Standard" Ground Truth.

CRITICAL RULES:
Both candidates will likely have a similar concise style. Your job is to identify which candidate provides a more complete, nuanced, and structurally sound critique.
1. NUANCE AND COMPLETENESS: Reward the candidate that successfully bridges general visual context with the specific technical critique requested, mirroring the depth of the Gold Standard.
2. FACTUAL ACCURACY: Penalize any candidate that hallucinates details not present in the Gold Standard.

You must output a strictly formatted JSON object with your step-by-step reasoning and your final decision ("A", "B", or "Tie").
\end{lstlisting}
\end{tcolorbox}
    \caption{LLM-as-judge system prompt for Solo vs. Shared Expert comparison.}
    \Description{Verbatim system prompt given to the LLM judge.}
    \label{fig:judge_solo_vs_shared}
\end{figure}

\begin{figure}[tbp]
    \centering
\begin{tcolorbox}[
    colback=gray!5,
    colframe=gray!30,
    width=\linewidth,
    boxrule=0.5pt,
    arc=0mm
]
\begin{lstlisting}[basicstyle=\scriptsize\ttfamily, breaklines=true, frame=none]
# EVALUATION WRAPPER PROMPT
User Prompt: "{prompt_text}"

Gold Standard (Ground Truth):
{ground_truth}

Candidate A:
{cand_a}

Candidate B:
{cand_b}

Evaluate which candidate better matches the Gold Standard in factual accuracy, style, and according to your specific rules. 
Output JSON strictly in this format:
{
    "reasoning": "<brief step-by-step explanation referencing your rules>",
    "winner": "<A, B, or Tie>"
}
\end{lstlisting}
\end{tcolorbox}
    \caption{The final evaluation wrapper prompt used to present candidates to the judge.}
    \Description{Verbatim system prompt given to the LLM judge.}
    \label{fig:judge_wrapper}
\end{figure}


\section{Additional Results}
\label{appen:additional_results}

This appendix collects supporting material for the results in the main text: the router confusion matrices discussed in Section~\ref{sec:res-router}, the visualization of the disentanglement ablation summarized in Table~\ref{tab:entanglement_ablation}.

\begin{figure}[tbp]
    \centering
    \includegraphics[width=0.49\columnwidth]{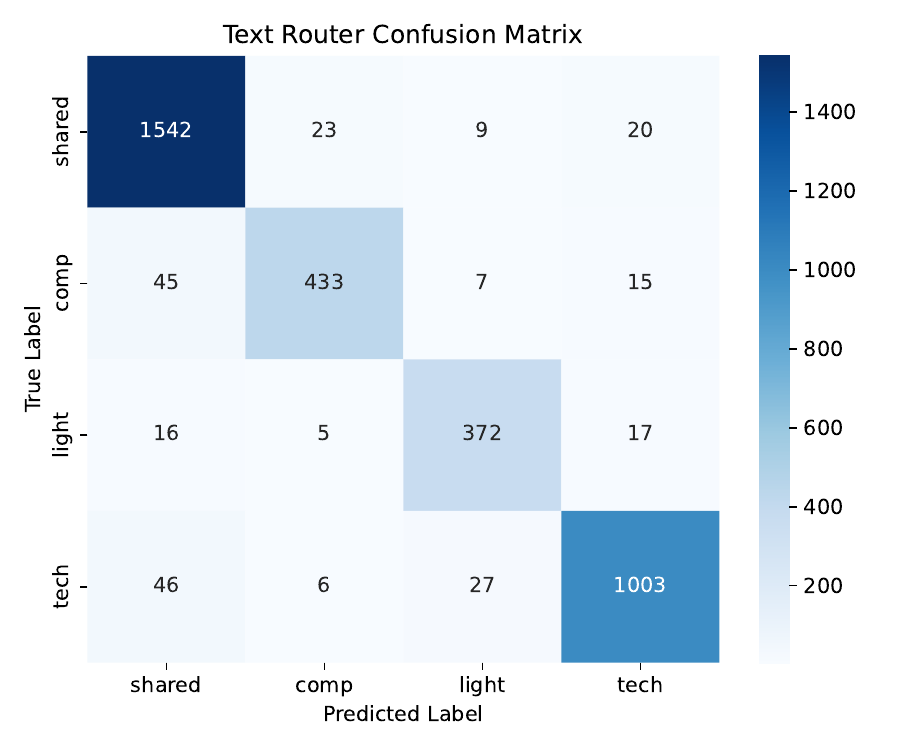}\hfill
    \includegraphics[width=0.49\columnwidth]{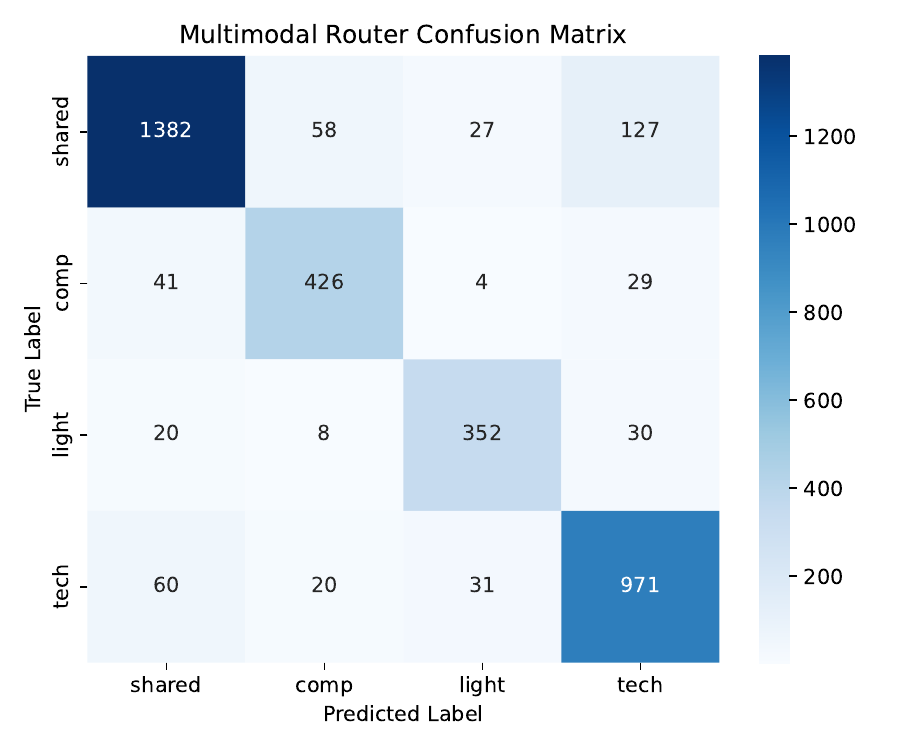}
    \caption{Confusion matrix for the text-only (left) and multimodal (right) router.}
    \Description{Two four-by-four confusion matrices over the shared, composition, lighting, and technical routing classes for the text-only router and the multimodal router.}
    \label{fig:text_router_cm}
\end{figure}

\begin{figure}[tbp]
    \centering
    \includegraphics[width=\columnwidth]{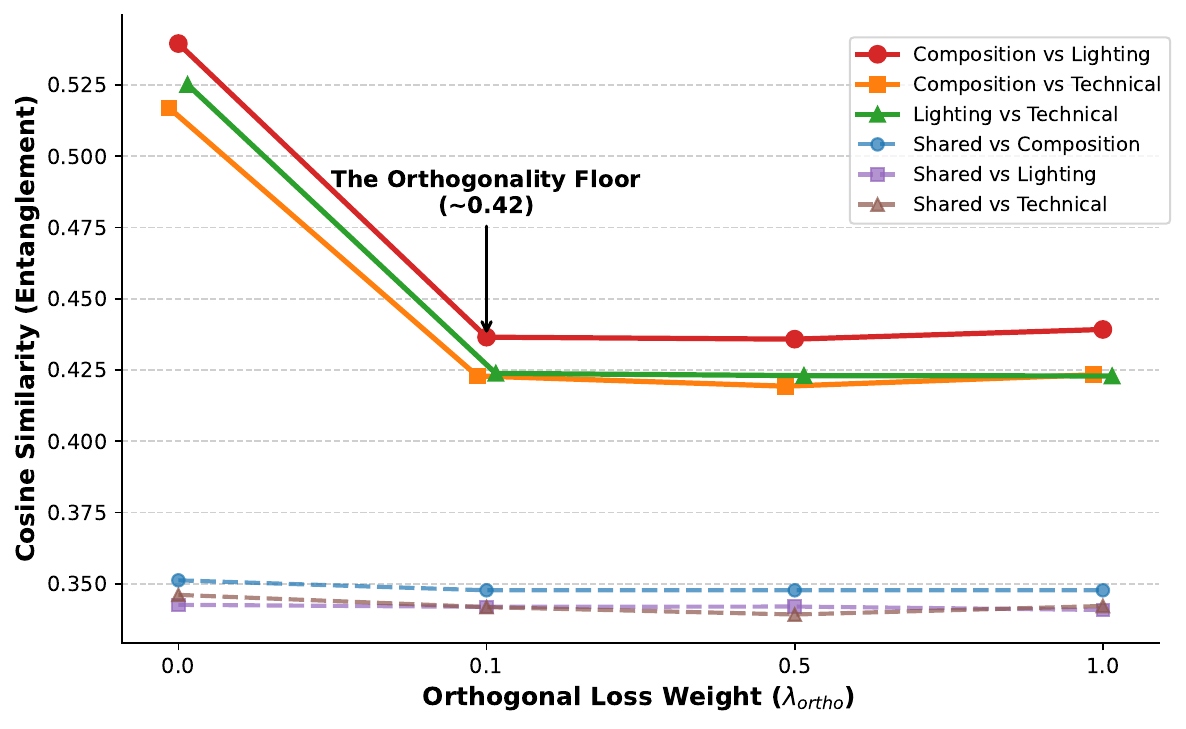}
    \caption{SVD-based expert disentanglement across different $\lambda_{\text{ortho}}$ constraints.}
    \Description{Line chart of pairwise SVD cosine similarity between expert adapters as the orthogonal penalty increases from 0 to 1.}
    \label{fig:disentanglement_test}
\end{figure}

\fi

\end{document}